\documentclass[conference]{IEEEtran}
\usepackage[T1]{fontenc}
\usepackage{tikz}

\usepackage{float}

\usepackage{algorithm}
\usepackage{algorithmic}
\usepackage{graphicx}

\usepackage{amsmath,amssymb,amsfonts}
\usepackage{soul}

\usepackage{tabularx}
\usepackage{multirow}
\usepackage{dcolumn}
\usepackage{booktabs}

\usepackage{subcaption}
\usepackage{enumerate}
\usepackage{subfig}
\usepackage{fancyhdr}
\usepackage{rotating} 

\usepackage{pgfplots}
\pgfplotsset{compat=1.18}

\usepackage{verbatim}
\usetikzlibrary{patterns}

\graphicspath{{./images/}}

\usepackage{nomencl}
\makenomenclature
\usepackage{colortbl}

\definecolor{Gray}{gray}{0.83}

\newcolumntype{x}{>{\columncolor{Gray}}c}
\newcolumntype{y}{>{\columncolor{white}}c}

\usepackage{etoolbox}
\renewcommand\nomgroup[1]{%
  \item[\bfseries
  \ifstrequal{#1}{A}{Acronyms}{%
  \ifstrequal{#1}{S}{Symbols}{%
  \ifstrequal{#1}{U}{Subscripts and Superscripts}{}}}%
]}

\usepackage{tikz}

\begin{document}

\title{CoBA: Integrated Deep Learning Model for Reliable Low-Altitude UAV Classification in mmWave Radio Networks\\

\vspace{-2mm}
}

\author{\IEEEauthorblockN{Junaid Sajid\IEEEauthorrefmark{1},
Ivo Müürsepp\IEEEauthorrefmark{1},
Luca Reggiani\IEEEauthorrefmark{2},
Davide Scazzoli\IEEEauthorrefmark{2},
Federico Francesco Luigi Mariani\IEEEauthorrefmark{2},
\\ Maurizio Magarini\IEEEauthorrefmark{2},
Rizwan Ahmad\IEEEauthorrefmark{3}, and
Muhammad Mahtab Alam\IEEEauthorrefmark{1}}

\IEEEauthorblockA{\IEEEauthorrefmark{1}Thomas Johann Seebeck Department of Electronics, Tallinn University of Technology (TalTech), Estonia}
\IEEEauthorblockA{\IEEEauthorrefmark{2}Politecnico di Milano - polimi, Milano, Italy}
\IEEEauthorblockA{\IEEEauthorrefmark{3}National University of Sciences and Technology (NUST), Islamabad, Pakistan}
\IEEEauthorrefmark{1}\{junaid.sajid, ivo.müürsepp, muhammad.alam\}@taltech.ee, \IEEEauthorrefmark{2}\{luca.reggiani, davide.scazzoli, federicofrancesco.mariani, \\ maurizio.magarini\}@polimi.it, \IEEEauthorrefmark{3} rizwan.ahmad@seecs.edu.pk
\vspace{-5mm}

}

\newcommand{\authnotice}{%
This paper has been accepted for publication at the IEEE International Conference on Communications (ICC) 2026.
This is the author’s version of the manuscript; the final version will appear in IEEE Xplore.}

\pagestyle{fancy}
\fancyhf{}

\fancyhead[C]{\authnotice}

\fancyfoot[C]{%
\footnotesize
\authnotice\\
\thepage
}

\renewcommand{\headrulewidth}{0pt}
\renewcommand{\footrulewidth}{0pt}

\maketitle
\thispagestyle{fancy} 

\begin{abstract}

Uncrewed Aerial Vehicles (UAVs) are increasingly used in civilian and industrial applications, making secure low-altitude operations crucial. In dense mmWave environments, accurately classifying low-altitude UAVs as either inside authorized or restricted airspaces remains challenging, requiring models that handle complex propagation and signal variability. This paper proposes a deep learning model, referred to as CoBA, which stands for integrated Convolutional Neural Network (CNN), Bidirectional Long Short-Term Memory (BiLSTM), and Attention which leverages Fifth Generation (5G) millimeter-wave (mmWave) radio measurements to classify UAV operations in authorized and restricted airspaces at low altitude. The proposed CoBA model integrates convolutional, bidirectional recurrent, and attention layers to capture both spatial and temporal patterns in UAV radio measurements. To validate the model, a dedicated dataset is collected using the 5G mmWave network at TalTech, with controlled low altitude UAV flights in authorized and restricted scenarios. The model is evaluated against conventional ML models and a fingerprinting-based benchmark. Experimental results show that CoBA achieves superior accuracy, significantly outperforming all baseline models and demonstrating its potential for reliable and regulated UAV airspace monitoring.
\end{abstract}


\vspace{-2mm}

\section{Introduction}
\label{sec:intro}
Uncrewed Aerial Vehicles (UAVs), commonly known as drones, have become integral to modern technological ecosystems, supporting diverse applications in surveillance, logistics, environmental monitoring, and smart city infrastructures~\cite{wandelt2023aerial, xu2023survey}. Their versatility, cost efficiency, and autonomous operation make them essential in the civilian and industrial sectors. However, this broad adoption raises major challenges in airspace management, in ensuring that UAVs remain within authorized airspaces and avoid sensitive regions. Unauthorized UAV activities in restricted airspace complicate airspace management and classification, potentially leading to security violations, privacy intrusions, and operational hazards near critical infrastructure. Therefore, effective monitoring and classification methods are increasingly needed to reliably identify UAV activities in both authorized and restricted airspace under complex and dynamic conditions~\cite{khan2024security}. 
UAVs naturally interact with cellular infrastructures, enabling the use of communication signals to detect and classify UAVs. These radio signals capture the propagation and interference characteristics of UAV links and can serve as valuable features for airspace monitoring and UAV detection~\cite{khawaja2025survey}. 
5G New Radio (NR) systems operating in the sub-6 GHz range have been widely used for UAV monitoring due to their broad coverage and stable connectivity. In contrast, mmWave bands offer higher bandwidth and capacity, promising improved UAV detection and classification~\cite{hong2021role}.


Several studies have investigated UAV detection using cellular networks, radio measurements, and Machine Learning (ML). In \cite{posch2023classifier}, the 5G NR based approach was proposed for height classification of mobile devices using Reference Signal Received Powe (RSRP), where Decision Tree (DT) achieves 98\% accuracy at all heights, while K-Nearest Neighbors (KNN) records 77\% at 40 m, 94\% at 80 m, and 95\% at 200 m. In~\cite{sheikh2019drone}, radio features such as  Received Signal Strength Indicator (RSSI), Signal-to-Interference-plus-Noise Ratio (SINR), and cell overlap were used to classify UAVs and regular user entities in mobile networks. Detection accuracy exceeded $99\%$ above $60\,$m but dropped sharply at lower heights, where DT achieved $54.9\%$ and $92\%$, and K-Nearest Neighbors (KNN) $62.4\%$ and $91.3\%$ at $15$ and $30\,$m, respectively. Similarly,~\cite{ryden2019rogue} evaluated Logistic Regression (LR) and DT models using RSRP, showing effective high-altitude rogue drones in mobile networks but only $5\%$ and $40\%$ accuracy at $15$ and $30\,$m.
Finally,~\cite{solomitckii2018technologies} showed the potential of 5G mmWave systems, where dense network deployment and directional antennas help in accurately classifying UAVs.

Existing methods perform well in detecting UAVs at medium to high altitudes (above 50 m AGL). However, achieving reliable classification at low altitudes (below 50 m AGL) remains a challenge, as conventional machine learning models such as DT, LR, and KNN~\cite{sun2024advancing} experience a marked decline in performance~\cite{posch2023classifier, sheikh2019drone, ryden2019rogue}. This limitation stems from complex propagation effects such as multipath fading, signal reflections, and interference at low altitude, which introduce nonlinear distortions in radio features such as RSRP, RSSI, and SINR. Conventional ML algorithms are limited by linear or distance-based decision rules and cannot adequately handle these nonlinear variations or account for the temporal patterns in UAV movement. These limitations highlight the need for models that capture the nonlinear and temporal patterns of low-altitude UAV signals. Furthermore, most existing studies focus on sub-6 GHz networks, while mmWave networks are rarely explored. In contrast, mmWave’s higher bandwidth and directional propagation characteristics offer greater potential to enhance UAV detection and classification at low altitudes.

To address these challenges, this study provides a unique contribution by focusing on mmWave-based low-altitude UAV classification through a deep learning (DL) model termed CoBA, which integrates a Convolutional Neural Network (CNN), a Bidirectional Long Short-Term Memory (BiLSTM), and an Attention mechanism. CNN extracts high-level spatial features from input signals~\cite{ige2024state}, BiLSTM captures bidirectional temporal dependencies~\cite{siami2019performance}, and the Attention layer emphasizes the most informative patterns~\cite{brauwers2021general}. Collectively, these components enable CoBA to model complex, dynamic, and nonlinear channel variations more effectively, addressing the limitations of traditional ML models and improving low-altitude UAV classification accuracy. In summary, the main contributions of this work are as follows.
\begin{itemize}
    \item \textbf{CoBA DL architecture:} CoBA integrates CNN, BiLSTM, and Attention within a unified model specifically designed for reliable low-altitude UAV classification;
    \item \textbf{Real-world 5G mmWave dataset:} UAV flights conducted at TalTech collected 5G mmWave measurements, including Physical Cell Identity (PCI), Synchronization Signal Block Index (SSB Idx), RSSI, SSB-RSSI, the SS-RSRP, the SS-SINR, and SS-RSRQ, providing a novel benchmark for airspace monitoring. To the best of our knowledge, this is the first use of measurements in this frequency band for UAV classification;
    \item \textbf{Extensive performance assessment:} CoBA is comprehensively evaluated against baseline ML models and a fingerprinting (FP) benchmark, demonstrating superior classification accuracy and robustness.
\end{itemize}

\vspace{-4mm}

\section{System and Signal Model}
\label{sec:system_model}
We consider a 5G mmWave-based UAV monitoring system where multiple radio units (RUs) provide coverage over the observation area. The objective of the system is to classify whether the UAV operates within an authorized airspace or has deviated into a restricted airspace, based on the measured radio parameters~\cite{morais2024key,ghosh20195g}. 
Let the set of RUs be defined as
\begin{equation}
\mathcal{R} = \{1, 2, \ldots, r, \ldots, R\},
\end{equation}
where each RU $r \in \mathcal{R}$ transmits reference signals at power $P_r$ on mmWave carriers. The position of the UAV $\mathbf{p}(t)$ in a given time instant $t$ is given by
\begin{equation}
\mathbf{p}(t) = [x(t), y(t), h(t)],
\end{equation}
where $x(t)$, $y(t)$, and $h(t)$ are longitude, latitude, and altitude, respectively. The UAV receives signals subject to large-scale path loss, shadowing, and small-scale fading~\cite{bai2014coverage}.

\subsection{Received Signal Model}
The \textit{complex baseband signal} received by the UAV corresponding to the $i$th reference signal resource element (RE) on the $b$th SSB at the time instant $t$ from the $r$-th RU can be expressed as
\begin{equation}
y^{(i)}_{r,b}(t) = \sqrt{P^{(i)}_{r,b}} h^{(i)}_{r,b}(t) s^{(i)}_{r,b}(t) + n^{(i)}(t),
\end{equation}
where $P^{(i)}_{r,b}$ is the transmitted power, $h^{(i)}_{r,b}(t)$ is the complex channel coefficient, assuming a non-frequency selective channel, $s^{(i)}_{r,b}(t)$ is the known transmitted symbol with normalized unit power, and $n^{(i)}(t) \sim \mathcal{CN}(0,\sigma^2)$ is the complex additive white Gaussian noise with zero mean and variance $\sigma^2$~\cite{ghosh20195g}. 
The \textbf{RSRP} from the $r$-th RU is given by
\begin{equation}
\rho_{r,b}(t) = \frac{1}{N_{RS}} \sum_{i=1}^{N_{RS}} \left| y_{r,b}^{(i)}(t) \right|^2,
\end{equation}
where $N_{RS}$ is the number of reference resource elements used in the average and $y_{r,b}^{(i)}(t)$ the signal in the time domain corresponding to the $i$-th reference resource element~\cite{morais2024key}. Similarly, \textbf{RSSI} is given by
\begin{equation}
\gamma_{r,b}(t) =\rho_{r,b}(t)+I_{r,b}(t),
\end{equation}
where 
\begin{align}
I_{r,b}(t) &= \frac{1}{N_{RS}} \sum_{i=1}^{N_{RS}} \left|\sum_{\substack{r' = 1 \\ r' \neq r}}^{R} \sqrt{P^{(i)}_{r',b}} h^{(i)}_{r',b}(t) s^{(i)}_{r',b}(t)\right|^2\nonumber \\ +&\frac{2}{N_{RS}} \sum_{i=1}^{N_{RS}}\Re \hspace{-.1cm} \left\{{y_{r,b}^{(i)}}^\ast\hspace{-.1cm}(t) \hspace{-.1cm} \sum_{\substack{r' = 1 \\ r' \neq r}}^{R} \sqrt{P^{(i)}_{r',b}} h^{(i)}_{r',b}(t) s^{(i)}_{r',b}(t) \right\}
\end{align} 
is the in-band interference from non-serving RUs~\cite{morais2024key}, with $^\ast$ denoting the complex conjugate. 
The \textbf{SINR} is given by
\begin{equation}
\zeta_{r,b}(t) = \frac{\rho_{r,b}(t)}{ I_{r,b}(t) + \sigma^2}.
\end{equation}

Finally, \textbf{RSRQ} is given by
\begin{equation}
\theta_{r,b}(t) = \frac{N_{RB} \cdot \rho_{r,b}(t)}{\gamma_{r,B}(t)},
\end{equation}
where $N_{RB}$ is the number of resource blocks over which RSSI is measured~\cite{ghosh20195g}.
  
\subsection{UAV Classification Model}
The UAV classification task is formulated as a supervised DL problem, where the goal is to distinguish UAV operations in authorized and restricted airspace based on 5G mmWave radio measurements. Let 
\begin{equation}
\label{eq9}
\mathbf{z}_b(t) = \big[ \rho_{1,b}(t), \dots, \rho_{R,b}(t), \ \gamma_{r,b}(t), \ \zeta_{r,b}(t), \ \theta_{r,b}(t)\big],
\end{equation}
be the measurement feature vector at time $t$.
A DL classifier $f(\cdot)$, represented by the proposed CoBA model, is a function that learns to map the input feature vector $\mathbf{z}(t)$ to the corresponding class label $\hat{y}(t)$:
\begin{equation}
\hat{y}_b(t) = f(\mathbf{z}_b(t)) \in \{0,1\},
\end{equation}
where $\hat{y}_b(t)=0$ and $\hat{y}_b(t)=1$ denotes UAV operation in authorized airspace and restricted airspace respectively.  



\section{Proposed DL Model for UAV Classification}
\begin{figure}[!b]
    \centering
    \includegraphics[width=\linewidth]{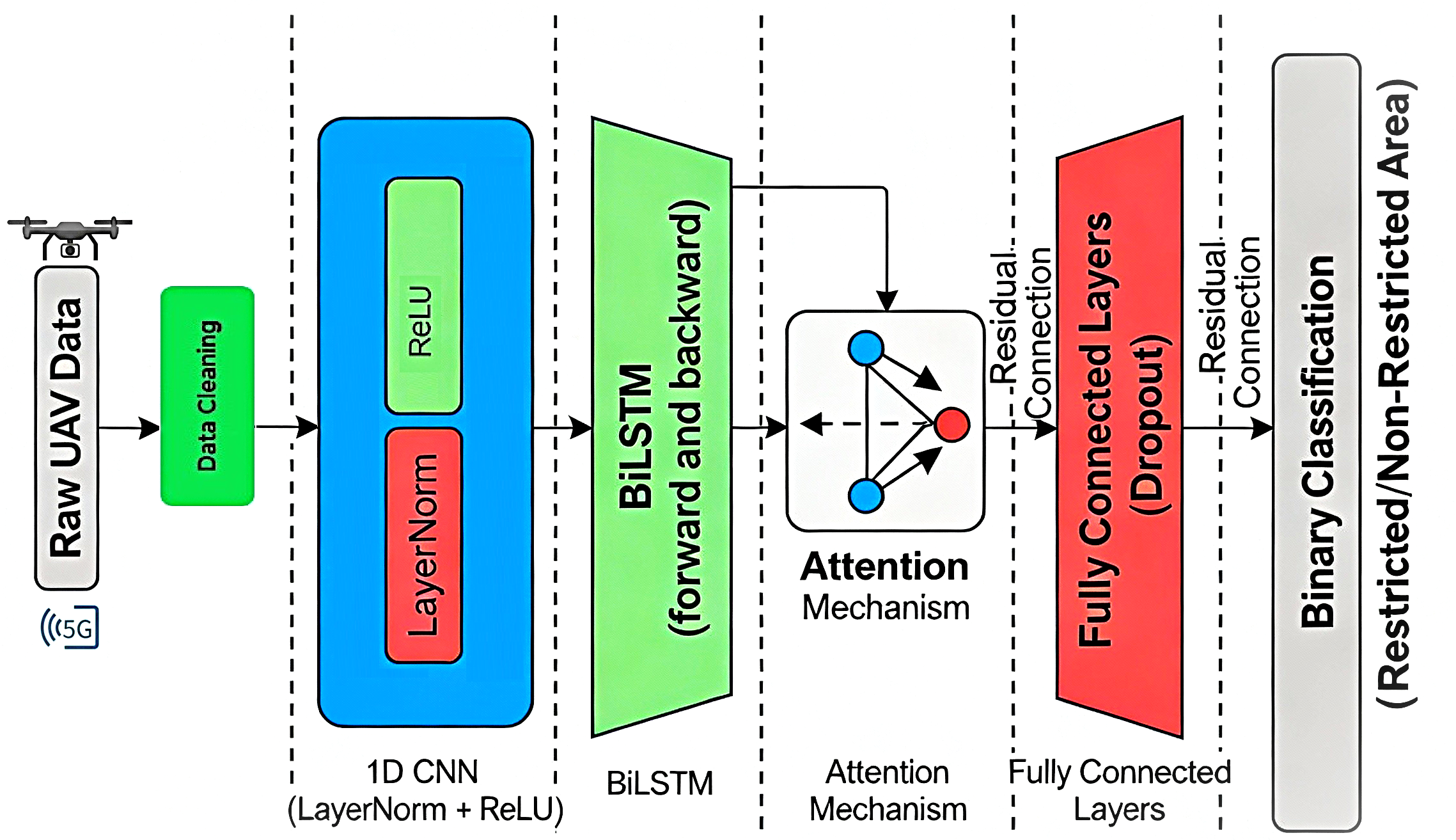}
    \caption{System Architectur}
    \label{fig:arch}
    \vspace{-6mm}
\end{figure}

This section presents the design and implementation of the CoBA DL model for classifying UAV operations into restricted and non-restricted airspace using 5G mmWave measurements. The model integrates CNN, BiLSTM, an attention mechanism, and fully connected layers with a residual connection to capture both spatial and temporal dependencies in the data. The system architecture is illustrated in Fig.~\ref{fig:arch}, with detailed components explained below.

\subsubsection{Model Initialization}
The model is initialized with four main components: a 1D CNN with two sequential convolutional layers with Layer Normalization and ReLU activation extract local patterns from input sequences of length $l$ with $f$ features per time step.; a BiLSTM that Processes sequences in both forward ($\overrightarrow{\mathbf{h}}_f$) and backward ($\overleftarrow{\mathbf{h}}_b$) directions to capture long-term temporal dependencies; an attention mechanism that computes attention weights ($\alpha_t$) for each time step to produce a context vector ($\mathbf{c}$) that emphasizes the most informative measurements; and fully connected layers that refines the context vector and adds a residual link from the attention output to the final logits ($\hat{\mathbf{y}}$), improving the training stability.

\begin{algorithm}[!b]
\caption{CoBA model for UAVs classification}
\label{alg:CoBA_detailed}
\footnotesize
\begin{algorithmic}[1]
\REQUIRE Raw UAV dataset $d = \{x, y\}$, sequence length $l$, number of features $f$ (see Eq.\ref{eq9}) , batch size $b$, CNN channels $c$, LSTM hidden size $h$, number of epochs $e$
\ENSURE Trained CoBA model and evaluation metrics

\STATE \textbf{Set seed:} Fix random seeds for reproducibility (PyTorch, NumPy, CUDA)
\STATE \textbf{Data splitting:} Stratify-split $d$ into 70\% train, 15\% val, 15\% test
\STATE \textbf{Sequence generation:}
\FOR{each set $s$ $in$ $\{train, val, test\}$}
    \FOR{$i = 1$ to $|s|-l+1$}
        \STATE $\mathbf{x}_i = [x_i, x_{i+1}, \dots, x_{i+l-1}] \in \mathbb{R}^{l \times f}$
        \STATE $y_i = y_{i+l-1}$
    \ENDFOR
\ENDFOR
\STATE \textbf{Dataloaders:} Mini-batch size $b$, use WeightedRandomSampler for training

\STATE \textbf{Initialize CoBA model:} CNN + BiLSTM + Attention + Fully Connected + Residual

\STATE \textbf{Forward pass (for input sequence $\mathbf{x}_i$):}
    \STATE \textbf{1D CNN: Local feature extraction}
    \STATE \hspace{1em} $\mathbf{h}^{(1)} = \text{ReLU}\big(\text{LayerNorm}(\mathbf{x}_i * \mathbf{w}^{(1)} + \mathbf{b}^{(1)})\big)$
    \STATE \hspace{1em} $\mathbf{h}^{(2)} = \text{ReLU}\big(\text{LayerNorm}(\mathbf{h}^{(1)} * \mathbf{w}^{(2)} + \mathbf{b}^{(2)})\big)$ \\ \hspace{1em}
    \COMMENT{$\mathbf{h}^{(2)} \in \mathbb{R}^{l \times c}$}

\STATE \textbf{BiLSTM: Temporal dependencies}
    \STATE \hspace{1em} $\overrightarrow{\mathbf{h}}_f \gets \text{LSTM}_f(\mathbf{h}^{(2)}_t)$ \COMMENT{Forward LSTM}
    \STATE \hspace{1em} $\overleftarrow{\mathbf{h}}_b \gets \text{LSTM}_b(\mathbf{h}^{(2)}_t)$ \COMMENT{Backward LSTM}
    \STATE \hspace{1em} $\mathbf{h}^{\text{lstm}}_t \gets [\overrightarrow{\mathbf{h}}_f; \overleftarrow{\mathbf{h}}_b]$
\COMMENT{$\mathbf{h}^{\text{lstm}} \in \mathbb{R}^{l \times 2h}$}

\STATE \textbf{Attention mechanism:}
\FOR{$t = 1$ to $l$}
    \STATE $e_t = \mathbf{h}^{\text{lstm}}_t \mathbf{w}_a + b_a$
    \STATE $\alpha_t = \dfrac{\exp(e_t)}{\sum_{k=1}^{l} \exp(e_k)}$
\ENDFOR
\STATE $\mathbf{c} = \sum_{t=1}^{l} \alpha_t \mathbf{h}^{\text{lstm}}_t$
\COMMENT{Context vector $\mathbf{c} \in \mathbb{R}^{2h}$}

\STATE \textbf{Fully connected + residual:}
    \STATE \hspace{1em} $\mathbf{z}_1 = \text{ReLU}(\mathbf{w}_1 \mathbf{c} + \mathbf{b}_1)$
    \STATE \hspace{1em} $\mathbf{z}_2 = \text{Dropout}(\mathbf{z}_1)$
    \STATE \hspace{1em} $\hat{\mathbf{y}} = \mathbf{w}_2 \mathbf{z}_2 + \mathbf{b}_2 + (\mathbf{w}_r \mathbf{c} + \mathbf{b}_r)$
    \STATE \hspace{1em} $\mathbf{p} = \text{Softmax}(\hat{\mathbf{y}})$
    \STATE \hspace{1em} $\hat{y} = \arg\max_{i} p_i$
\STATE \textbf{Loss function:} Weighted cross-entropy loss
    \STATE \hspace{1em} $l = - \sum_{i} w_i y_i \log(p_i)$
\STATE \textbf{Optimizer:} AdamW with learning rate $\eta = 10^{-4}$

\FOR{epoch = 1 to $e$}
    \STATE \textbf{Training:} Forward pass $\rightarrow$ compute $l$ $\rightarrow$ backward pass $\rightarrow$ update weights $\rightarrow$ gradient clipping
    \STATE \textbf{Evaluation:} Compute loss, accuracy, precision, recall, F1-score
\ENDFOR
\STATE \textbf{Output:} Trained model parameters and evaluation metrics
\end{algorithmic}
\end{algorithm}

\subsubsection{Forward Pass}
For each input sequence $\mathbf{x}_i$, the following steps are performed: The sequence passes through two CNN layers: the first layer computes $\mathbf{h}^{(1)}$ $=$ $\text{ReLU}$$(\text{LayerNorm}$$(\mathbf{x}_i * \mathbf{w}^{(1)}$ $+$ $\mathbf{b}^{(1)}))$, all the w and b terms represent learnable parameters (weights and biases). The second layer produces $\mathbf{h}^{(2)}$ $=$ $\text{ReLU}$$(\text{LayerNorm}$$(\mathbf{h}^{(1)} * \mathbf{w}^{(2)} + \mathbf{b}^{(2)})))$, which has shape $l \times c$, encoding short-term temporal patterns in radio features such as RSSI, RSRP, SINR, and RSRQ. The CNN output is passed through a BiLSTM: the forward LSTM generates $\overrightarrow{\mathbf{h}}_f$, and the backward LSTM generates $\overleftarrow{\mathbf{h}}_b$. The hidden states are concatenated as $\mathbf{h}^{\text{lstm}}_t = [\overrightarrow{\mathbf{h}}_f; \overleftarrow{\mathbf{h}}_b]$ with shape $l \times 2h$, capturing long-term dependencies along the UAV trajectory.

Attention scores $e_t$ are calculated for each time step and normalized to weights $\alpha_t$ using softmax. $\mathbf{c} = \sum_{t=1}^{l} \alpha_t \mathbf{h}^{\text{lstm}}_t$ is he context vector which aggregates the most relevant information from the sequence, focusing on decisive measurements. The context vector is passed through fully connected layers with ReLU and dropout to yield $\mathbf{z}_2$, with a residual connection adding the original vector to the logits:
\begin{equation}
\hat{\mathbf{y}} = \mathbf{w}_2 \mathbf{z}_2 + \mathbf{b}_2 + (\mathbf{w}_r \mathbf{c} + \mathbf{b}_r)
\end{equation}
The final class probabilities $\mathbf{p}$ are obtained via softmax, and the predicted class $\hat{y}$ corresponds to the highest probability.

\subsubsection{Loss Function and Optimization}
The CoBA model employs weighted cross-entropy loss to handle class imbalance, where $w_i$ represents the weight of each class. Optimization is performed using AdamW with a learning rate of $10^{-4}$, which adapts the learning rates per parameter and includes weight decay to prevent overfitting.

\subsubsection{Training Procedure}
The model is trained over $e$ epochs with batch size $b$. Each epoch consists of: (i) Forward pass to compute predictions; (ii) Loss computation using weighted cross-entropy; (iii) Backward pass to calculate gradients; (iv) Weight update using AdamW; (v) Gradient clipping to maintain stability. After each epoch, the model is evaluated using accuracy, precision, recall, and F1 score for a comprehensive assessment. The training and inference procedure of the CoBA model is summarized in Algorithm~\ref{alg:CoBA_detailed}.

\section{Experimental Testbed for Low-Altitude UAV Classification in mmWave Radio Networks}
This section describes the experimental setup and the data collection procedure for UAV classification using the Tallinn University of Technology 5G mmWave network. It presents details on the deployment of RUs, UAV instrumentation, flight campaigns, and the preparation of the resulting dataset for the DL model.

\subsection{Measurement Setup}
\begin{figure}[t]
    \centering
    \includegraphics[width=\linewidth]{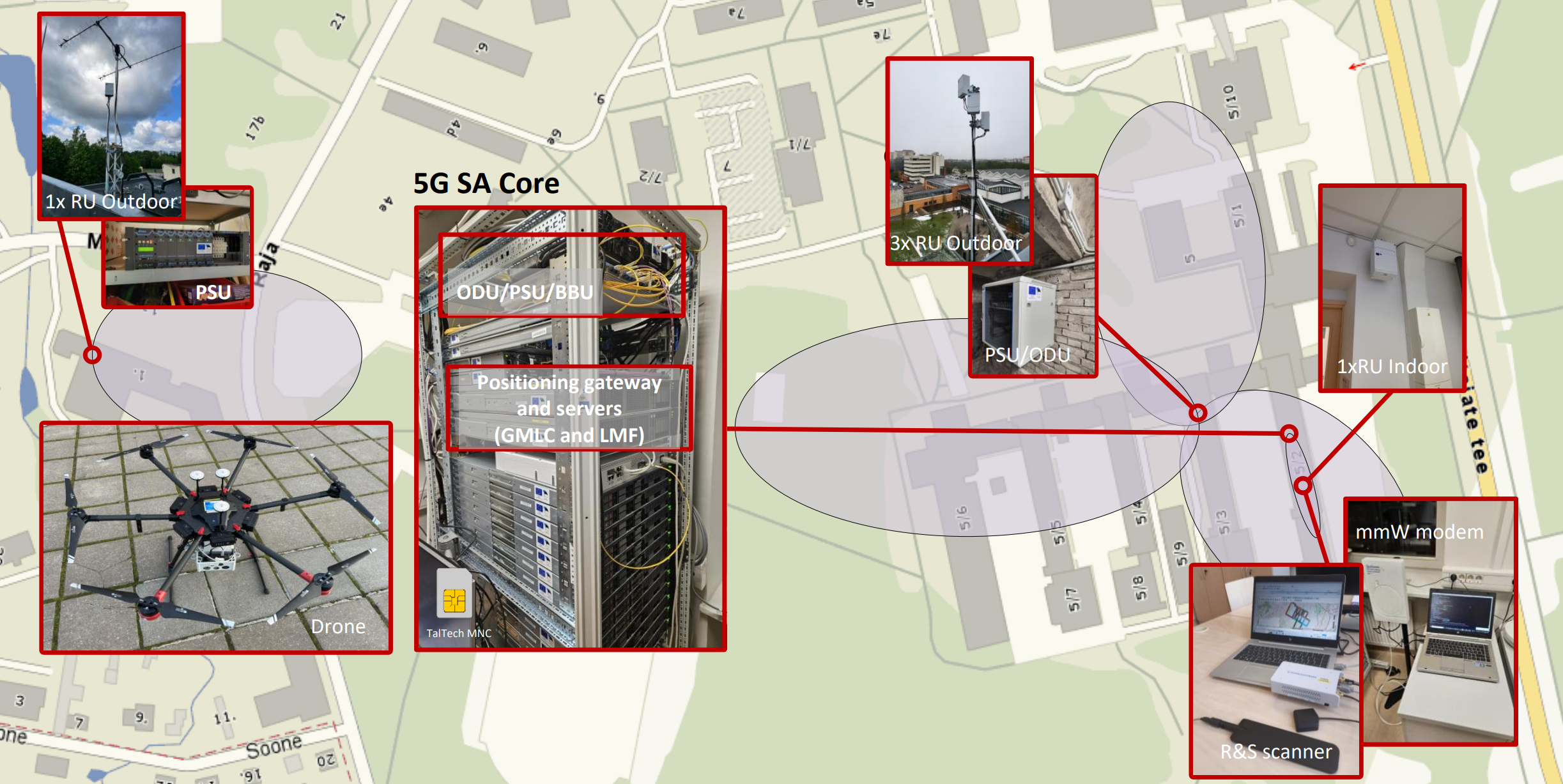}
    \caption{5G Advanced mmWave setup at TalTech featuring RUs, BBU, Ericsson 5G SA core, UAV, and radio scanner}
    \label{fig:taltech}
\end{figure}

The measurement campaigns are conducted at the Tallinn University of Technology by using 5G Advanced campus testbed, which features mmWave radio network and 5G SA infrastructure operating in the n258 band under an Estonian spectrum license. The UAV is equipped with a vertically polarized omnidirectional antenna (QMS-00029) covering $24-40\,$GHz, with an estimated gain of $3\,$dBi. The setup includes four outdoor and one indoor RU connected to a centralized baseband unit, ensuring seamless campus-wide coverage. The parameters of three outdoor RUs used in this study are given in Table \ref{tbl:RUs}. UAV has mmWave radio scanner mounted to collect all the radio measurements data being used for the classification task. This study focuses on classifying UAVs operating in restricted or authorized airspace (at low-altitude),  with the experimental setup illustrated in Fig.~\ref{fig:taltech}. 

\vspace{-2mm}

\begin{table}[!h]
\centering
\caption{Outdoor RUs Deployment Configuration}
\label{tbl:RUs}
\begin{tabular}{ll}
\hline
\textbf{Parameter} & \textbf{Description} \\ \hline
RU Model & Ericsson AIR 5322 \\ \hline
Maximum EIRP & $62\,$dBm (configuration-dependent) \\ \hline
Operating Frequencies & 24.353760 GHz and 24.422880 GHz \\ \hline
Total Down-Tilt & $0^\circ$ \\ \hline
Deployment Location & Taltech Campus  \\ \hline
Coordinates & $59.394834^\circ$ N, $24.670878^\circ$ E \\ \hline
\textbf{RU 0} & Bearing: $26^\circ$, Height: $20.73\,$m AGL\\ \hline
\textbf{RU 1} & Bearing: $292^\circ$, Height: $20.55\,$m AGL\\ \hline
\textbf{RU 2} & Bearing: $116^\circ$, Height: $20.24\,$m AGL \\ \hline
\end{tabular}
\end{table}


\subsection{Data Measurement Campaigns}
Data measurement campaigns are carried out using multiple UAV flights in both restricted and non-restricted airspace. The airspace located very close to the Tallinn University of Technology campus building, where the three outdoor RUs shown in Fig.~\ref{fig:taltech} are installed, is considered a restricted airspace, while the open airspace away from these buildings is considered an allowed airspace. All flights are carried out at altitudes of $50\,$m or below.
Consequently, we categorized these as low-altitude flights. A DJI Matrice 600 Pro is used for all flights, maintaining a speed of $1$ to $2\,$m/s. An iPad Mini 5 running the DJI GS Pro application is used to manage automated flight paths. All measurements are carried out under stable weather conditions, limited to sunny or cloudy skies, with no rain or snow, and wind speeds below $8\,$m/s within the flight area. During the flights, UAV's onboard Rohde \& Schwarz (R\&S) TSMx6 drive and walk test scanner, together with R\&S®ROMES4 software, records 5G NR signal measurements.

\subsection{Dataset Preparation}
The scanner is configured to export raw measurement logs in text file. Subsequently, these files are converted to the CSV format for further analysis. Samples from take-off and landing are removed, leaving only in-flight measurements. The resulting dataset contains 58,788 samples. The raw radio measurement data consists of a wide range of physical layer indicators captured per SSB beam. Each detected cell is represented by up to $32$ storage slots, denoted as \texttt{Mbr} 1 – 32, per frequency channel.
For each sample, a structured following features are recorded: \textbf{PCI}, \textbf{SSB Idx}, \textbf{RSSI}, \textbf{SSB-RSSI}, \textbf{SS-RSRP}, \textbf{SS-SINR}, and \textbf{SS-RSRQ}. A cleaning procedure is applied to harmonize column names, remove redundant prefixes.
To enable supervised learning, a binary label (\texttt{class}) is added, where \texttt{class~0} represents UAVs in non-restricted airspace, and \texttt{class~1} indicates restricted sirspace.
Two labeled datasets are merged into one, with missing values handled by linear interpolation followed by median imputation to reduce bias. Data is split into two stratified: $70\%$ training, $15\%$ validation, and $15\%$ testing to preserve class balance. To capture short-term temporal variations, the UAV flight data is divided into overlapping windows of length $L=10$, each labeled by its final sample.

\section{Results and Evaluation}
\label{sec:results}
This section presents the analysis of the features and complexity of the models. The experiments are conducted on a system equipped with an Intel(R) Core(TM) Ultra 5 125U processor ($1.30$ GHz) and $32$ GB RAM. Deep learning models are implemented in Python using Jupyter Notebook, and R\&S scanner software is also operated on the same system. Our model is analyzed and compared with ML models used in literature studies~\cite{posch2023classifier, sheikh2019drone, ryden2019rogue} such as SVM, KNN, DT, and LR, as well as a DL model LSTM and a non-ML FP model to evaluate its performance.

The FP approach, commonly used in positioning tasks, involving: (i) partitioning the flight area into 3D elements with dimensions $10 [m] \times 10 [m] \times 10 [m]$; (ii) assigning each element a class label (authorized or restricted) based on measurements within it, similar to labels in the CoBA model; (iii) using 70\% of the data to build the 3D labeled map (training) and the rest for testing; (iv) classifying new points as authorized or restricted based on the minimum Euclidean distance to labeled elements.
The FP approach is assumed to be a benchmark for ML models and further support to the identification and explanation of the parameters that have a dominant impact in the performance.

\begin{table}[b]
\renewcommand{\arraystretch}{0.7} 
\setlength{\tabcolsep}{11pt} 
\centering
\caption{Performance Comparison of Different Models using All Features}
\label{tbl:model_comparison}
\begin{tabular}{lcccc}
\toprule
\textbf{Model} & \textbf{Accuracy} & \textbf{Precision} & \textbf{Recall} & \textbf{F1 Score} \\
\midrule
SVM & 0.9035 & 0.8446 & 0.9960 & 0.9140 \\
KNN & 0.9009 & 1.0000 & 0.8075 & 0.8935 \\
DT & 0.9143 & 0.8574 & 1.0000 & 0.9232 \\
LR & 0.9140 & 0.8570 & 1.0000 & 0.9230 \\
\midrule
FP & 0.9978 & 1.0000 &  0.9950 & 0.9978 \\
\midrule
LSTM & 0.9129 & 0.8562 & 0.9987 & 0.9220 \\
\textbf{CoBA} & \textbf{0.9989} & \textbf{0.9989} & \textbf{0.9989} & \textbf{0.9989} \\
\bottomrule
\vspace{-5mm}
\end{tabular}
\end{table}
\begin{table}[b]
\renewcommand{\arraystretch}{0.6} 
\setlength{\tabcolsep}{4pt} 
\centering
\caption{Results After Training Individual Parameters}
\label{tab:individual_results}
\begin{tabular}{lcccccc}
\toprule
\textbf{Model} & \textbf{PCI} & \textbf{SSB Idx} & \textbf{RSSI} & \textbf{SS-RSRP} & \textbf{SS-SINR} & \textbf{SS-RSRQ} \\
\midrule
SVM  & 0.9143 & 0.9142 & 0.9014 & 0.8770 & 0.8356 & 0.8584 \\
KNN  & 0.8981 & 0.8945 & 0.9013 & 0.9013 & 0.8934 & 0.8978 \\
DT   & 0.9143 & 0.9143 & 0.9013 & 0.8921 & 0.8758 & 0.8820 \\
LR   & 0.9057 & 0.8796 & 0.7625 & 0.7917 & 0.7224 & 0.7714 \\
\midrule
FP   & 0.9664 & 0.6822 & 0.7780 & 0.6923 & 0.6587 & 0.6569 \\
\midrule
LSTM & 0.9143 & 0.9124 & 0.5151 & 0.6883 & 0.8677 & 0.8304 \\
\textbf{CoBA} & \textbf{0.9973} & \textbf{0.9881} & \textbf{0.8156} & \textbf{0.8573} & \textbf{0.8783} & \textbf{0.8397} \\
\bottomrule
\end{tabular}
\end{table}

 \vspace{-2mm}

\subsection{Feature Analysis}
\begin{figure}[!t]
    \centering
    \includegraphics[width=\linewidth]{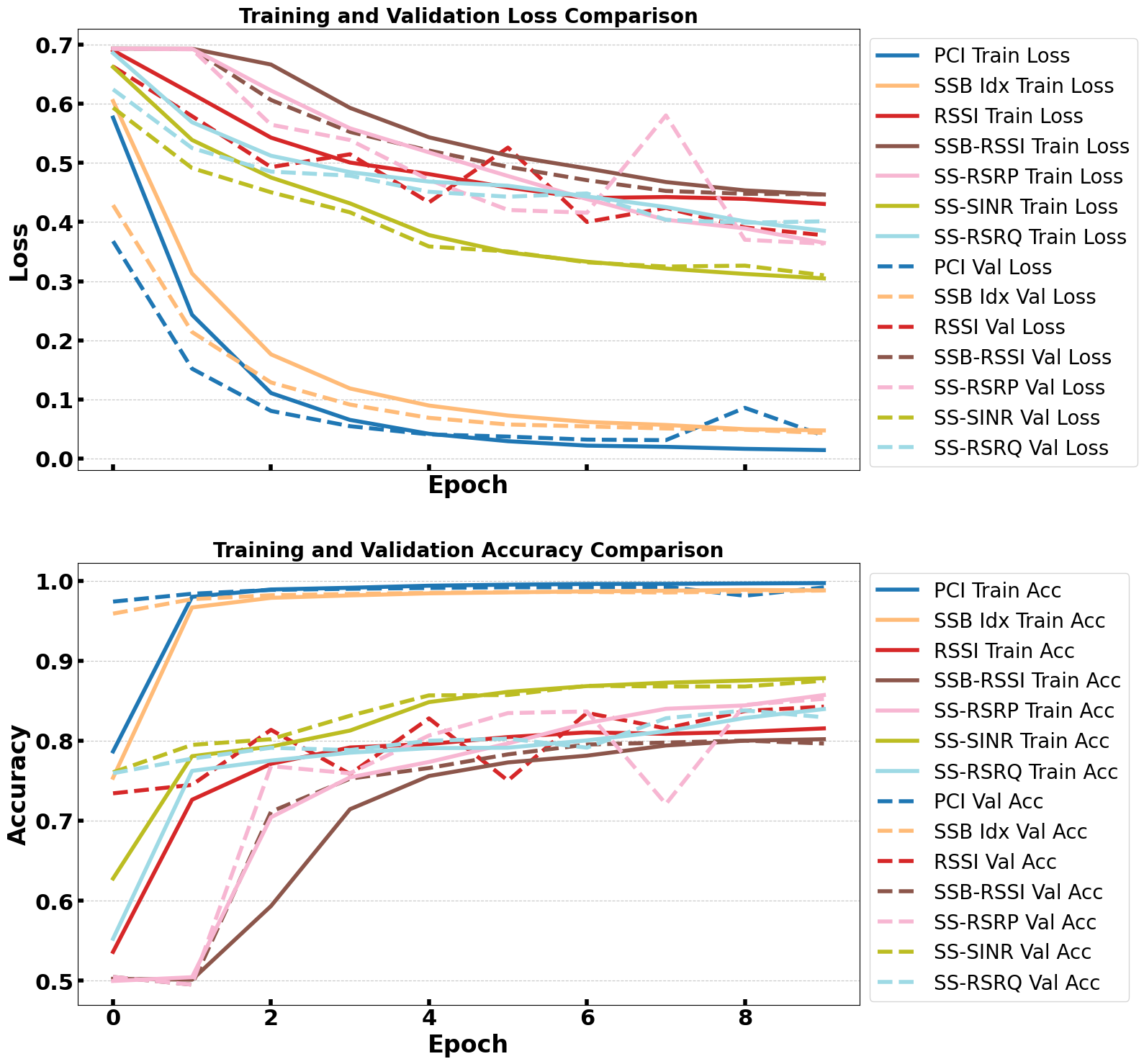}
    \caption{Performance of the CoBA with individual features.}
    \label{fig:loss_accuracy}
\end{figure}

The feature analysis confirms that the proposed CoBA model demonstrates exceptional reliability and accuracy in low-altitude 5G mmWave environments. As shown in Table~\ref{tbl:model_comparison}, when trained on all parameters, the CoBA model achieved $99.89\%$ accuracy, outperforming baseline models (around $90$–$91\%$) and the FP ($99.78\%$). Individual feature evaluation (Table~\ref{tab:individual_results} and Fig.~\ref{fig:loss_accuracy}) shows that PCI and SSB Idx are the most decisive, with the CoBA model achieving accuracy $99.73\%$ and $98.81\%$ correspondingly, and FP achieving $96.73\%$ and $68.81\%$, respectively, while other power-related features yielded moderate results between $81$–$87\%$. This is primarily because PCI and SSB Idx provide spatial and beam-specific identifiers that remain stable across varying channel conditions, effectively capturing the topological relationship between the UAV and serving cells. In contrast, power-related parameters such as RSRP, RSRQ, and SINR are more sensitive to multipath fading, obstruction, and small-scale variations in low-altitude environments, which reduces their decisive capability for reliable airspace classification.

\begin{table}[t]
\renewcommand{\arraystretch}{0.7} 
\setlength{\tabcolsep}{11pt} 
\centering
\caption{Performance Comparison of Different Models using PCI and SSB Idx}
\label{tbl:model_comparison_two_features}
\begin{tabular}{lcccc}
\toprule
\textbf{Model} & \textbf{Accuracy} & \textbf{Precision} & \textbf{Recall} & \textbf{F1 Score} \\
\midrule
SVM                & 0.9119 & 0.9993 & 0.8226 & 0.9024 \\
KNN                & 0.9030 & 1.0000 & 0.8039 & 0.8913 \\
DT       & 0.9122 & 1.0000 & 0.8226 & 0.9027 \\
LR & 0.9095 & 0.8453 & 1.0000 & 0.9162 \\
\midrule
FP   & 0.9970 & 1.0000 & 0.9942 & 0.9971
 \\
\midrule
LSTM & 0.9095 & 0.8453 & 1.0000 & 0.9162 \\
\textbf{CoBA} & \textbf{0.9982} & \textbf{0.9982} & \textbf{0.9982} & \textbf{0.9982} \\
\bottomrule
\end{tabular}
\end{table}

\begin{figure}[!t]
    \centering
    \includegraphics[width=\linewidth]{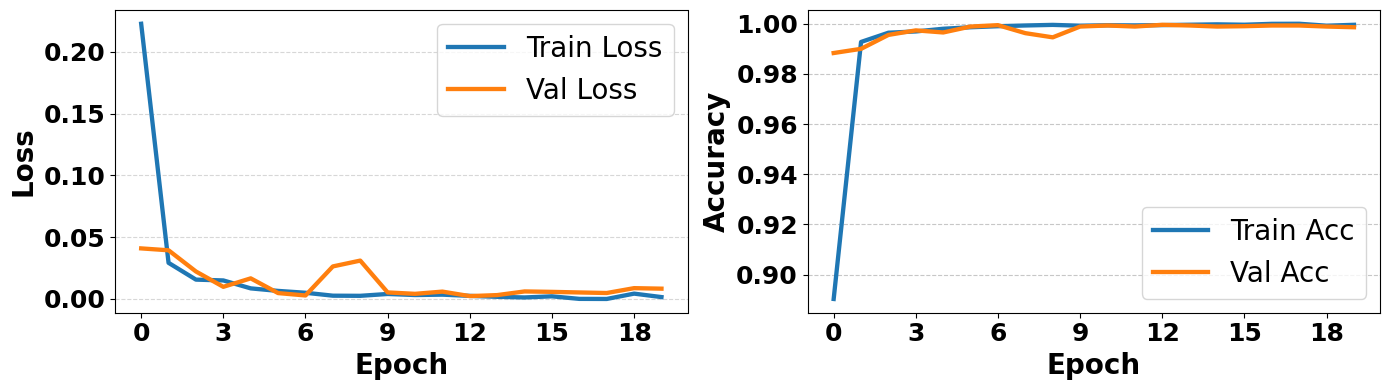}
    \caption{Performace of the CoBA using PCI and SSB Idx}
    \label{fig:CoBA_loss_acc}
    \vspace{-5mm}
\end{figure}

Since PCI and SSB Idx are identified as the most decisive features, further analysis is performed by training the CoBA model using only these two features. Fig.~\ref{fig:CoBA_loss_acc} illustrates the accuracy and loss over iterations: in the early stages, the validation performance fluctuates, but both the accuracy and the loss stabilize as the training progresses, indicating robust convergence. The results in Table~\ref{tbl:model_comparison_two_features} demonstrate that using only PCI and SSB Idx, the CoBA model achieves 99.82\% accuracy that closely matches the performance obtained using all features. This shows that these two parameters are the dominant beam-related factors that influence the performance of the model.

Given that the model’s performance with all features and with only PCI and SSB Idx differs by less than 0.1\%, these parameters clearly have the greatest impact on the learning and prediction capability. This finding introduces an important insight: \textbf{in mmWave communication at low altitudes, PCI and SSB Idx play a more decisive role in UAV classification tasks than power-related metrics}. This aspect, confirmed by the FP results (Tables \ref{tbl:model_comparison}-\ref{tbl:model_comparison_two_features}), has not been explicitly addressed in prior studies, which mainly focus on power-related features for UAV detection and classification~\cite{posch2023classifier, sheikh2019drone, ryden2019rogue}.


\subsection{Complexity Analysis}

\begin{figure}[!t]
    \centering
    \includegraphics[width=\linewidth]{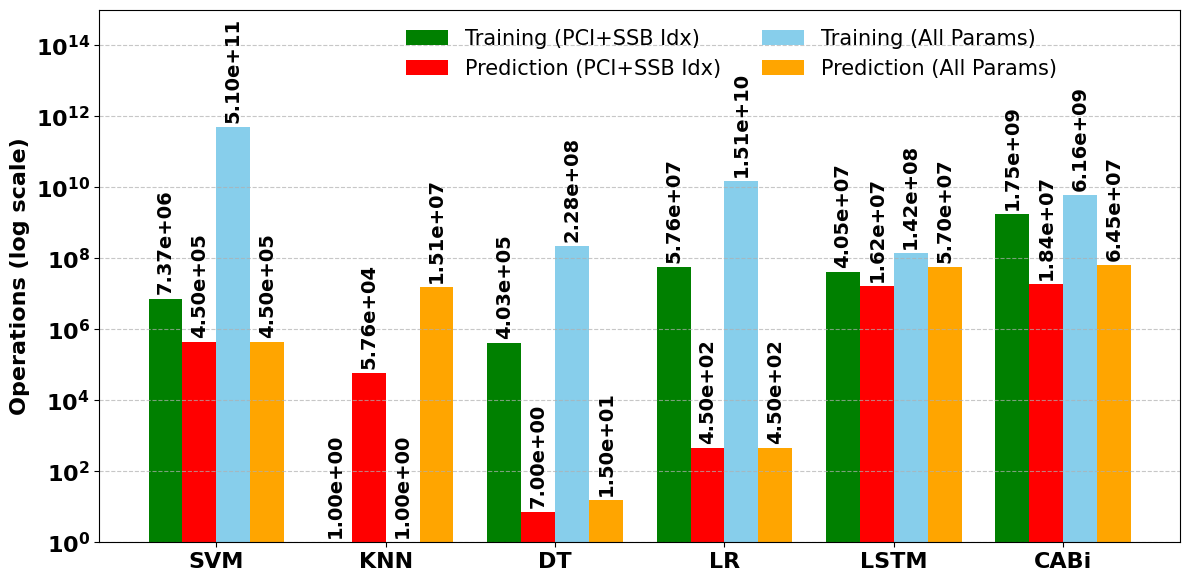}
    \caption{Algorithmic Complexity of the Models}
    \label{fig:complexity}
    \vspace{-5mm}
\end{figure}

Figure~\ref{fig:complexity} presents a comparative analysis of algorithmic complexity for models showing the number of operations (on logarithmic scale) for the training and prediction phases under two configurations: using only PCI and SSB Idx, and using all features. The results show that while the CoBA model exhibits the highest initial complexity even with two features, its computational cost increases modestly approximately 2.6 times in training and 2.4 in prediction when moving from the reduced to the full feature set. In contrast, the baseline ML models such as DT, and SVM show a dramatic rise in complexity, exceeding 200–500 times for training. Despite this, CoBA maintains nearly the same accuracy while achieving far better scalability with increasing feature dimensions. This shows the reliability of our model with increasing features and highlights a favorable trade-off between performance and computational cost. The complexity of the FP model is not calculated because 70\% of the data is used to build area maps and only 30\% for testing; therefore, FP serves solely as a benchmark for reference.


\section{Conclusion}
\label{sec:conclusion}
To conclude, this study presents a significant advancement in the domain of UAV monitoring and classification within 5G mmWave communication environments. The proposed CoBA model, which integrates CNN, BiLSTM, and an attention mechanism, establishes a highly accurate and reliable model to distinguish UAV operations in restricted and authorized airspaces. Not only does it achieve exceptional classification performance that exceeds accuracy $99\%$  but also maintains robustness when relying solely on two dominant features, PCI and SSB Index, demonstrating computational efficiency and feature significance. This capability enables reliable airspace monitoring and rapid detection of restricted UAV activity, which supports safer integration of drones into regulated environments. As UAVs become increasingly vital to industrial and civilian applications, the proposed model contributes to ensuring their responsible and secure deployment. Future work targets multiclass classification for ground users and UAVs at low and high altitudes, and deploying CoBA in real-time edge environments to enable scalable and secure airspace management.

\vspace{-5mm}

\section*{Acknowledgment}
This work is carried out and funded under the NATO SPS project G7699 – 'Passive Radar Observation and Detection of UAVs via Cellular Networks (PROTECT)'.

\bibliographystyle{ieeetr}
\bibliography{references}

\end{document}